\documentclass[letterpaper]{article} % DO NOT CHANGE THIS
\usepackage{arxiv}  % DO NOT CHANGE THIS
\usepackage{times}  % DO NOT CHANGE THIS
\usepackage{helvet}  % DO NOT CHANGE THIS
\usepackage{courier}  % DO NOT CHANGE THIS
\usepackage[hyphens]{url}  % DO NOT CHANGE THIS
\usepackage{graphicx} % DO NOT CHANGE THIS
\usepackage{natbib}  % DO NOT CHANGE THIS AND DO NOT ADD ANY OPTIONS TO IT
\usepackage{caption} % DO NOT CHANGE THIS AND DO NOT ADD ANY OPTIONS TO IT
\usepackage{color}
\usepackage{tabularx}
\usepackage{booktabs}
\usepackage{graphicx}
\usepackage{xspace}
\usepackage{multirow}
\usepackage{mathtools}
\usepackage{rotating}
\usepackage{url}
\usepackage{tabulary}
\usepackage{multirow}
\usepackage{subfigure}
\usepackage{amsmath}
\usepackage{slashbox}
\usepackage{makecell}
\usepackage{bigstrut}
\usepackage{enumerate}
\usepackage{makecell}
\usepackage{url}
\usepackage{threeparttable}
\usepackage{caption}
\usepackage{subfigure}
\usepackage{float}
\usepackage{siunitx}
\usepackage{paralist}
\usepackage{xcolor,colortbl}
\usepackage{tabularx}
\usepackage{array}
\usepackage{threeparttable}
\usepackage{enumitem}
\usepackage{pifont}       % \ding{xx}
\usepackage{tikz}
\usepackage{sidecap}
\usetikzlibrary{positioning}
\usepackage{caption}

\definecolor{shadecolorP}{RGB}{144,238,144}

\definecolor{shadecolorN}{RGB}{240,248,255}

\definecolor{orange}{RGB}{149,72,12}

\definecolor{olive}{RGB}{9,84,26}

\definecolor{green}{RGB}{15,109,37}

\definecolor{blue}{RGB}{12,80,151}

\usepackage{algorithm}
\usepackage{algorithmic}

\usepackage{newfloat}
\usepackage{listings}
\DeclareCaptionStyle{ruled}{labelfont=normalfont,labelsep=colon,strut=off} % DO NOT CHANGE THIS
\floatstyle{ruled}
\newfloat{listing}{tb}{lst}{}
\floatname{listing}{Listing}
\makeatletter
\def\hlinew#1{%
	\noalign{\ifnum0=`}\fi\hrule \@height #1 \futurelet
	\reserved@a\@xhline}

\title{Modeling Human Responses to Multimodal AI Content}
\author{
Zhiqi Shen\textsuperscript{\rm 1}\equalcontrib, Shaojing Fan\textsuperscript{\rm 2}\equalcontrib, \\ Danni Xu\textsuperscript{\rm 2}, Terence Sim\textsuperscript{\rm 2}, Mohan Kankanhalli\textsuperscript{\rm 2}
}
\affiliations{
\textsuperscript{\rm 1} Salesforce Research Asia, \textsuperscript{\rm 2} National University of Singapore \\

zhiqi.shen@salesforce.com, \{dcsfs, terence.sim\}@nus.edu.sg, \\ dannixu@u.nus.edu, mohan@comp.nus.edu.sg
}

\usepackage{bibentry}
\begin{document}

\maketitle

\begin{abstract}

As AI-generated content becomes widespread, so does the risk of misinformation. While prior research has primarily focused on identifying whether content is authentic, much less is known about how such content influences human perception and behavior. In domains like trading or the stock market, predicting how people react (\emph{e.g.}, whether a news post will go viral), can be more critical than verifying its factual accuracy. To address this, we take a human-centered approach and introduce the MhAIM Dataset, which contains 154,552 online posts (111,153 of them AI-generated), enabling large-scale analysis of how people respond to AI-generated content. Our human study reveals that people are better at identifying AI content when posts include both text and visuals, particularly when inconsistencies exist between the two. We propose three new metrics: \emph{trustworthiness}, \emph{impact}, and \emph{openness}, to quantify how users judge and engage with online content. We present \textbf{T-Lens}, an LLM-based agent system designed to answer user queries by incorporating predicted human responses to multimodal information. At its core is \textbf{HR-MCP} (Human Response Model Context Protocol), built on the standardized Model Context Protocol (MCP), enabling seamless integration with any LLM. This integration allows T-Lens to better align with human reactions, enhancing both interpretability and interaction capabilities. Our work provides empirical insights and practical tools to equip LLMs with human-awareness capabilities. By highlighting the complex interplay among AI, human cognition, and information reception, our findings suggest actionable strategies for mitigating the risks of AI-driven misinformation.

\end{abstract}

\section{1 \: Introduction}

\vspace{0.35cm}
%\quote
\emph{``Success in creating AI could be the biggest event in the history of our civilization. Or the worst.''}
\vspace{0.15cm}

\hspace*{\fill}\emph{---Stephen Hawking (2016)}\\
%\endquote 

As artificial intelligence (AI) technique rapidly advances, AI-generated content (AIGC) is becoming increasingly common in fields like journalism, art, and social media \cite{solaiman2023evaluating,du2023exploring,epstein2023art,cao2023comprehensive,masood2023deepfakes}. While the benefits of AIGC are manifold, the potential harms cannot be overlooked. For instance, AI-crafted content, especially when multimodal in nature, can often be indistinguishable from genuine human-crafted content, leading to potential misinformation, deception, and erosion of trust in digital platforms \cite{jo2023promise,uzun2023chatgpt,budak2024misunderstanding}. 

\begin{figure}[ht]
	\centering
	\includegraphics[width=.47\textwidth]{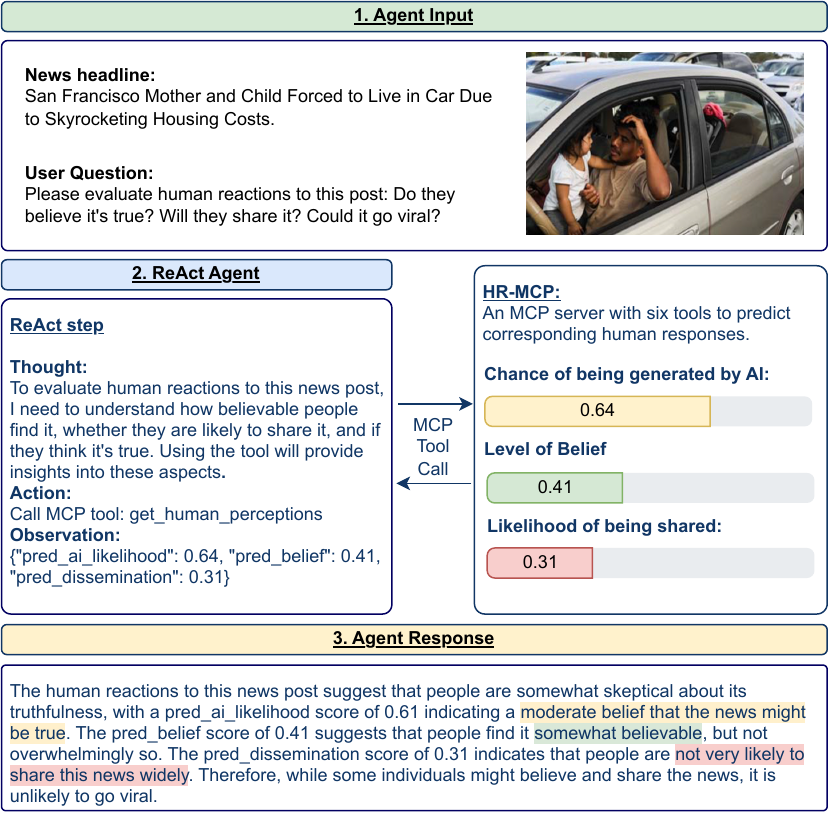}
	\caption{We study how users perceive AI-generated content and propose T-Lens, an LLM-based agent that predicts human responses with interpretable explanations. At its core is HR-MCP (Human Response Model Context Protocol), a module that reasons over both content and how humans are likely to interpret and respond to it.}%\vspace{-10pt}
	\label{fig:teaser}
\end{figure}

\begin{figure*}[ht]
	\centering
	\includegraphics[width=.99\textwidth]{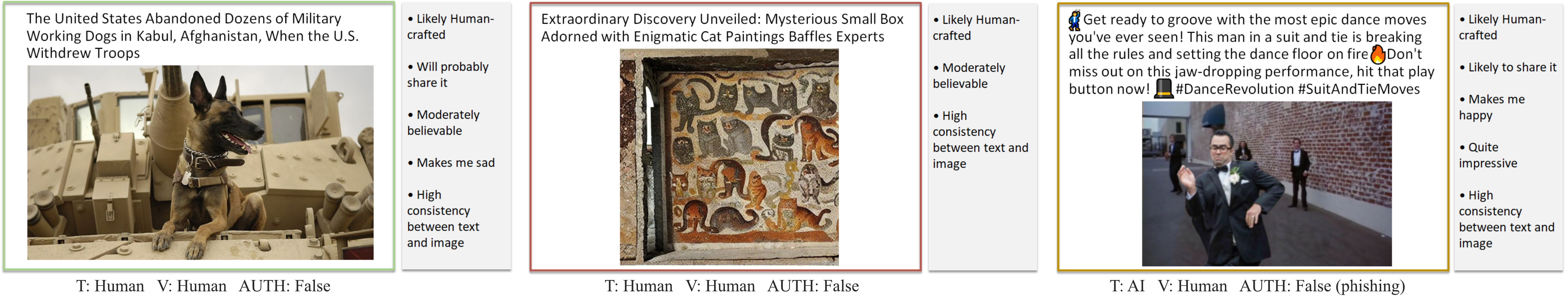}
	\caption{Examples from the MhAIM Dataset with origins of text (T) and visuals (V), authenticity (AUTH), and average human perceptions shown. Unless noted, AI texts and visuals are generated by ChatGPT and Stable Diffusion, respectively.}	
    \label{fig:example}
\end{figure*}

While a plethora of studies have addressed misinformation \cite{arora2021detecting,aimeur2023fake,he2023reinforcement} and AI-generated content \cite{malik2022deepfake,cao2023comprehensive,bandi2023power}, the current research primarily focuses on generating or detecting such content in isolation \cite{hartwig2024landscape,liu2024worldview}, or remains confined to a single modality \cite{mirsky2021creation,kreps2022all,stockl2023evaluating,zhou2023synthetic,fan2023advances}. There is still a big gap in understanding how people react to and are influenced by multimodal AI-generated content. Addressing this gap is essential for fostering trust in online platforms \cite{robertson2022potential}, advancing digital forensics \cite{gong2020deepfake}, and promoting the ethical use of AI \cite{hermann2022artificial,ferrara2024genai}.

To investigate the misinformation potential of AI-generated content (AIGC), we present \textbf{T-Lens} (Trust Lens), an agent system that predicts human responses to online content and provides accessible, interpretable explanations (Fig.~\ref{fig:teaser}). At its core is the \textbf{Human Response--Model Context Protocol (HR-MCP)}, a specialized module estimating human-centered attributes such as trustworthiness, impact, and openness. To train and evaluate T-Lens, we construct the \textbf{Multimodal Human \& AI Generated Media Dataset (MhAIM)}, comprising 154{,}552 posts produced by either humans or AI (see Fig.~\ref{fig:example} for examples). A human study on MhAIM reveals diverse reactions: AI-generated text is often seen as more trustworthy than AI visuals, and users are generally less inclined to believe or share content they suspect to be AI-generated. Yet, well-crafted AIGC can still be persuasive, even when its origin is recognized. These findings motivate the design of HR-MCP, which captures how people interpret and respond to multimodal content. Integrated into T-Lens, HR-MCP enhances transparency and helps mitigate the continued influence effect (CIE) of misinformation~\cite{ecker2022psychological,aslett2024online}. Experiments on real-world data underscore the effectiveness of our approach. A demo video and code are included in the supplementary material. %Upon acceptance, we will publicly release the dataset and code.

We adopt a human-centric approach to address the challenges posed by AIGC and misinformation. Rather than focusing solely on detection, we prioritize predicting how people respond to content. This perspective is particularly important in domains such as trading and the stock market, where anticipating user reactions (\emph{e.g.}, whether a news post will go viral or influence investment behavior), can be more consequential than verifying its factual accuracy~\cite{marquardt2025fakefin}. Our human studies show that well-crafted AIGC can still appear credible and persuasive, highlighting the limitations of relying solely on detection. In contrast to methods that depend on external fact-checking~\cite{qi2024sniffer,hangloo2023evidence}, which are increasingly challenged by sophisticated generative tools~\cite{wu2023ai,aslett2024online}, T-Lens aligns with human perception and offers a more robust alternative. Our findings offer valuable insights for trading platforms, news aggregators, and social media systems aiming to strengthen automated review and reduce the impact of AI-driven misinformation. Our contributes are as follows:

\begin{itemize}[leftmargin=*,noitemsep,topsep=3pt]
	
	\item We present \textbf{T-Lens}, an agent system designed to generate human-aligned responses to multimodal content. The key module of T-Lens is \textbf{HR-MCP}, a specialized MCP server developed based on insights from our human study. HR-MCP enables the prediction of human responses to misinformation, capturing aspects such as trustworthiness, impact, receptivity, user beliefs, and sharing tendencies. By integrating HR-MCP, T-Lens delivers accessible explanations that help mitigate the harmful effects of misinformation and address challenges posed by AIGC. Experimental results demonstrate that T-Lens significantly enhances the alignment of human responses produced by large language models. Furthermore, HR-MCP follows a modular MCP standard, enabling seamless integration with various LLMs for human-centric reasoning.

	\item We carefully designed the human study and found that user sensitivity to AIGC varies, peaking for multimodal social media posts and decreasing for text-only content. Despite general distrust, well-crafted AIGC can still be influential. These findings provide the community with insights to develop targeted strategies against AI-driven misinformation.	
	
	\item We construct MhAIM---a multimodal dataset of AI- and human-generated content. A subset of MhAIM has been annotated with human responses, including elicited emotions, perceived credibility, and the likelihood of sharing. MhAIM supports understanding human-AIGC interaction and aids in both countermeasures against malicious use of AIGC and strategies for its positive application.
	
\end{itemize}

\section{2 \: Related Work}

\textbf{Multimodal Analysis on Misinformation.} 
Multimodal analysis on misinformation leverages text, image, and video interactions \cite{von2020helping,zhou2020survey,aneja2021cosmos,sun2023inconsistent,shao2024dgm4++}, but this approach has not yet been commonly applied to AIGC. There is growing interest in detecting AI-generated content \cite{cao2023comprehensive,hu2023radar,du2023finguard,amoroso2023parents,lin2024detecting,zhu2024genimage}, using methods like deep learning for text \cite{chaka2023detecting} and neural networks for AI-synthesized imagery \cite{qi2023fakesv}, including deepfake videos \cite{mundra2023exposing} and real-time deepfakes \cite{mittal2022gotcha} (see review \cite{lin2024detecting}), however, they are mainly in single modality. We found well-crafted AIGC can still gain trust, so solely detecting it may not capture human responses adequately. Hence, we prioritize multimodal consistency, aligning with human perception and receptivity.

\noindent\textbf{Model Context Protocol.} 
The integration of external tools into large language models (LLMs) has evolved rapidly in recent years. OpenAI's introduction of the function calling mechanism in 2023 marked a pivotal step in this direction, allowing models to dynamically interface with external functions~\cite{anthropic2024mcp}. However, the function calling approach required manual interface definitions and authentication setups, limiting its scalability and general applicability across diverse use cases. Building upon this limitation, Anthropic introduced the Model Context Protocol (MCP) in late 2024, proposing a more unified and extensible framework for tool interoperability. MCP addresses key shortcomings of earlier approaches by offering a standardized communication protocol based on JSON-RPC 2.0, facilitating seamless integration across hosts, clients, and servers through structured STDIO and SSE channels \cite{SiloMCP}. By formalizing the roles of MCP hosts (AI applications), clients (connectors), and servers (capability providers), MCP significantly lowers the engineering overhead of tool integration and supports more robust, modular deployment of AI agents across various environments~\cite{hou2025model}. In this work, we extend the MCP framework to develop T-Lens, a modular agent system that integrates human-centered reasoning with dynamic tool interoperability, enabling robust prediction of human responses to multimodal content.

\begin{table}[h]\small
	\centering
	\begin{tabular}{p{3.2cm}p{4.7cm}}
		\hlinew{1pt}
		\textbf{Type} & \textbf{Detailed Attributes} \\
		\hline
		Emotional Responses & Happy? Sad? Angry? Frightened? Impressed? \\
		Behavioural Reactions & Will you share the content with friends? Do you believe this content? \\
		Perception Related to AI & Is this computer-generated? Why? \\
		Multimodal Consistency & Text matches the visual content? Textual sentiment matches the visual sentiment? Are objects mentioned in the text shown visually? \\
		Genre & Genre of the news/social media post? \\
		\hlinew{1pt}	
	\end{tabular}%
	\caption{Responses collected in our human study.}	
	\label{tab:annotation}%	
\end{table}%

\section{3 \: Dataset and Human Study}

We constructed the Multimodal Human \& AI Generated Media Dataset (MhAIM), which consists of \(111,153\) pieces of AI-generated content and \(43,309\) pieces of human-crafted materials (see examples in Fig.~\ref{fig:example}). 

\subsection{3.1 \: Stimuli Collection and AIGC Generation}

To build MhAIM, we integrated data from multimodal misinformation and authentic sources, including Fakeddit~\cite{nakamura2020r}, N24News~\cite{wang-EtAl:2022:LREC3}, FaceForensics++~\cite{rossler2019faceforensics++}, WildDeepfake~\cite{pu2021deepfake}, and Snopes~\cite{Snopes}, along with emotional datasets Emod~\cite{fan2018emotional} and NVED~\cite{fan2020and}. This diverse integration enables MhAIM to capture the complexity of both real and misleading content. See Supplementary Sec.1.1 for details.

We generated AIGC based on aforementioned real-world data. Text prompts from our collection were used with Stable Diffusion\cite{Stablediffusion} to produce images, and an AI Headline Generator, built using LangChain \cite{LangChain} and ChatGPT \cite{ChatGPT} with tools like VQA \cite{li2022blip,tong2022videomae}, produced diverse AI texts across styles and media types. We also used the Gemini API to create texts from original and AI-generated visuals. In total, we constructed 111,153 AI-generated entries. Details and examples are in Suppl. Sec. 1.1–1.2.

\subsection{3.2 \: Human Psychophysics Study}

We conducted an extensive human study using the MhAIM Dataset. We recruited 765 participants (ages between 19-65, 390 females) from the crowdsourcing platform Toluna \cite{Toluna}, spanning three countries (participant numbers in parentheses): the US (382), India (193), and Singapore (190). We further recruited 20 students (ages between 19-34, 12 females) from a local university. All participants received compensation for their involvement in the study. Participants viewed 30 posts (single or multimodal) and completed a questionnaire after each, assessing emotional response, behavioral tendencies (e.g., belief, sharing), perceived AI origin, and modality consistency (Table~\ref{tab:annotation}). Question phrasing varied by content type to capture how users perceive and engage with AIGC versus human content.

Due to budget constraints, our human study concentrated on a subset of MhAIM, encompassing $8,925$ posts from different origin (AI-generated or human-crafted) and type (news, social media posts and phishing messages). Each post, across different modalities of content, received annotations from 3 to 5 participants. Detailed study design, questionnaire, participants' demographics, and an overview of human-annotated data can be found in Suppl. Sec. 1.3-1.4. 

\subsection{3.3 \: Key Findings from Human Study}

\noindent \textbf{Definition and Method.} 
We assessed participants' ability to distinguish AI-generated from human-crafted content by asking them to rate the ``AIGC likelihood'' of each post. Following Signal Detection Theory (SDT)~\cite{wickens2001elementary}, we computed sensitivity \(d'\) and bias \(c'\), as used in prior misinformation studies~\cite{seo2019trust,batailler2022signal}. Sensitivity (\(d'\)) measures how well participants distinguish AI (``signal'') from human (``noise'') content, with higher values indicating better discrimination. Bias (\(c'\)) reflects response tendency: \(c'=0\) indicates no bias, \(c'>0\) favors human content, and \(c'<0\) favors AI content.

We employed inferential statistics to examine differences in the aforementioned metrics, using standard tests in behavioral sciences (see \cite{bailey2008design} for an overview). We evaluated whether each metric varied by multiple factors, initially using ANOVA to check for general effects, followed by post-hoc Tukey tests for significant ANOVAs to pinpoint manipulation effects, with a Bonferroni correction applied for multiple comparisons, setting a reduced alpha of $.01$. Below we report five major observations. Additional findings and detailed statistics with visualizations are reported in Suppl. Sec. 2.

\begin{figure}[ht]
	\centering
	\includegraphics[width=.47\textwidth]{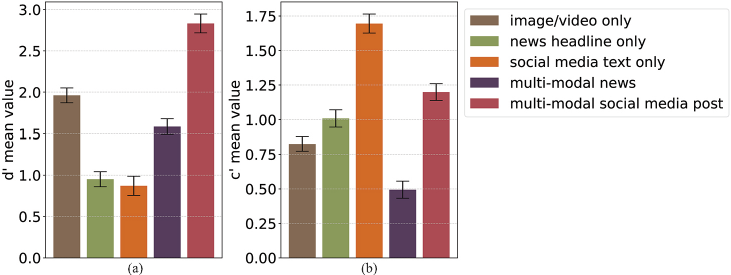}
	\caption{Users are most accurate at recognizing AI-generated multimodal social media posts and least accurate with AI-generated text (a). They tend to classify content as human-crafted, especially for social media posts, as indicated by the positive bias (\(c'\)) (b). Error bars represent standard errors, showing variations among individuals.}
	\vspace{-.1in}
	\label{fig:dprime} 
\end{figure}

\begin{table}[ht] \small
	\centering
	\begin{tabular}{p{0.095cm}<{\centering}|p{3.9cm}p{3.7cm}}
		\hlinew{1pt}
		& \textbf{Cues for AI origin} & \textbf{Cues for human origin} \\
		\hline
		\multirow{3}[2]{*}{\begin{sideways}News \hspace*{1.8em} \end{sideways}} & \textcolor{orange}{The news seems not true.} & \textcolor{orange}{The news seems true to me} \\
		& \textcolor{blue}{Texts and visuals mismatch in contents or sentiments.} & \textcolor{blue}{Texts and visuals align in content \& sentiment.} \\
		& The visual may be AI-made, but the text isn't. & The style is typical of news outlets \\
		\hline
		\multirow{3}[2]{*}{\begin{sideways} \fontsize{7.7}{2}\selectfont Social media post  \hspace*{0.001em} \end{sideways}} & \textcolor{blue}{Texts and visuals mismatch in contents or sentiments.} & \textcolor{blue}{Texts and visuals align in content \& sentiment.} \\
		& The content doesn't make sense & The style akin to common social media platforms \\
		& The visual may be AI-made, but the text isn't. & I have seen a similar post before \\
		\hlinew{1pt}    \end{tabular}%
	\caption{Participants' reasons for classifying multimodal information as AIGC or human-crafted. Content veracity (brown text) and text-visual mismatch (blue text) are top reasons for news and social media posts, respectively.}	
	\label{tab:reason}%	
\end{table}%

\begin{table*}[htbp]\small
	\centering
	\begin{threeparttable}[b]	
		\begin{tabular}{p{2cm}p{7.8cm}p{6.3cm}}
			\hlinew{1pt}
			\textbf{Metric} & \textbf{Definition} & \textbf{Measure}\\
			\hline
			Trustworthiness & The extent to which information elicits user trust & Belief likelihood $-$ AIGC likelihood \\
			Impact & Influence on individual and community & Belief likelihood + Dissemination propensity \\
			Openness & Users' willingness to accept and engage with perceived AIGC & (AIGC likelihood $+$ 1) $\times$ Impact \\			
			\hlinew{1pt}
		\end{tabular}%
	
	\end{threeparttable}
		\caption{Definition of three information metrics for the study of human receptivity towards AIGC. All attributes were normalized to [0, 1] using min-max normalization. }		
		\label{tab:openness}%		
\end{table*}%

\noindent \textbf{Human Sensitivity Towards AIGC.} \emph{Observation 1:} Participants' sensitivity towards AIGC varies based on modality and information type. Our participants exhibited the highest sensitivity for multimodal social media posts and the lowest for social media texts without accompanying visuals.

On average, participants did slightly better than random guessing (\(d'\) $ = 1.64 \pm 0.38$) in distinguishing AIGC from human-crafted content. The type of information they were judging made a big difference to their sensitivity, \(F(4, 5287) = 61.68, p < .00001\)\footnote{ANOVA results are reported as, ``$F$(df$_{condition}$, df$_{error}$) = $F$ value, $p$ = $p$ value'', with \(p < .05\) rejecting the null hypothesis.}. As shown in Fig.~\ref{fig:dprime} (a), our participants were best at spotting AI content when it included both text and visuals on social media (\(d' = 2.83\)), but had a harder time with text-only posts (\(d' = 0.87\)). Participants tended to be cautious during judgments, leaning towards classifying content as human-crafted (suggested by positive bias in Fig.~\ref{fig:dprime} (b)). 

\noindent \emph{Observation 2:} Mismatches between text and the corresponding visuals generally lead to the identification of posts as AI-generated. Specifically for news posts, the perceived veracity of the content plays a pivotal role.

As shown in Table \ref{tab:reason}, participants frequently identified multimodal content as AIGC due to mismatches between text and visuals. Specifically, for news content, perceived information credibility was a crucial factor.

\noindent \textbf{Receptivity and Trust in AIGC.} \emph{Observation 3:} AI text is more well-received than AI visuals, while human-crafted content, regardless of whether it's real or fake, consistently gets higher receptivity than AI-generated content.

Besides having participants assess whether a post was AI-generated (``AIGC likelihood''), we also gauged their level of belief in the content's veracity (``belief'') and their inclination to share it (``dissemination propensity''). Based on these ratings, we develop three novel metrics to evaluate human receptivity to AIGC\footnote{In our study, these metrics also apply to human-crafted content, as we present information to users without explicitly labeling it as AI or non-AI.} (Table \ref{tab:openness}): \emph{trustworthiness}, indicating user trust and calculated as the difference between belief and AIGC likelihood; \emph{impact}, measuring effect on individuals and the community, computed as the sum of belief and dissemination propensity; and \emph{Openness}, reflecting willingness to engage with perceived AIGC, computed by augmenting the product of AIGC likelihood and impact with the impact attribute. A post demonstrates high openness when users are prepared to believe and disseminate it, even if they think it is likely created by AI. 

Our human-crafted posts, whether real or fake, consistently exceeded AI-generated content in all three metrics. AI text significantly outperformed AI visuals in terms of openness \((F \geq 28.00, p \leq 0.0001)\), a disparity likely due to the more developed capabilities of AI in text generation compared to visual content creation. The openness of AIGC increased alongside stronger perceived positive sentiments \(( \rho \geq 0.60, p \leq 0.0001)\) and greater multimodal consistency \(( \rho \geq 0.54, p \leq 0.0001)\), underscoring the importance of these elements in influencing user acceptance. Visualizations are in Figs. S11-S12 in the supplementary.

\begin{figure}[ht]
	\centering
    \includegraphics[width=0.45\textwidth]
    {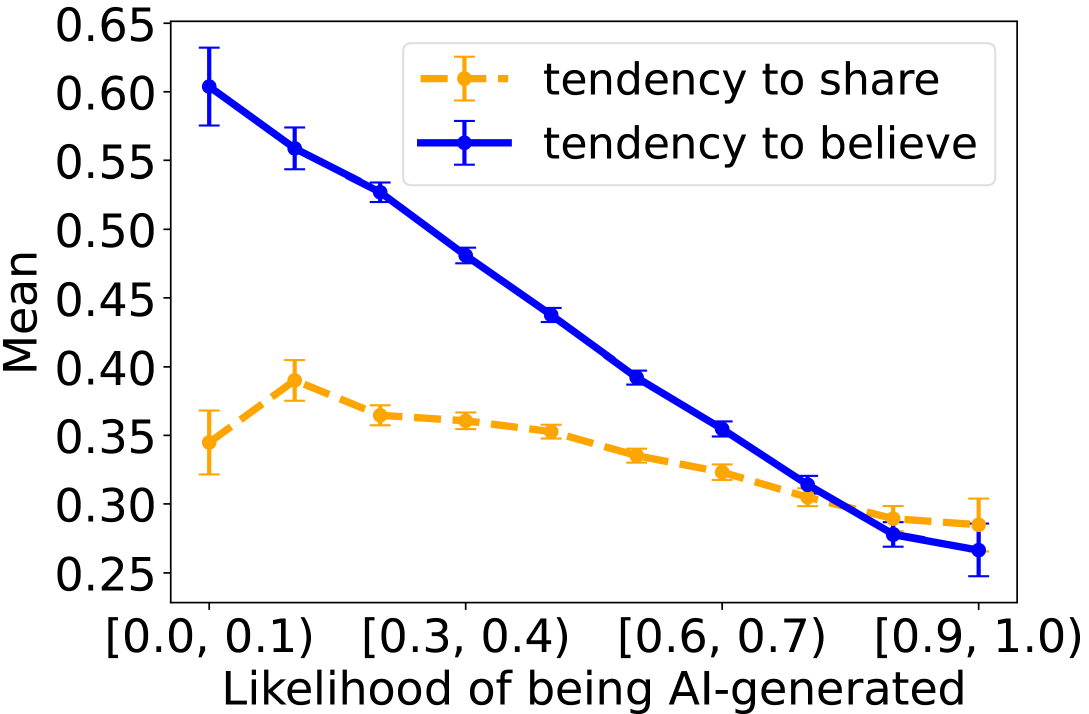}
	\caption{As participants grew more convinced a post was AI-generated, they were less willing to trust and share it.} %\vspace{-.15in}
\label{fig:sharing_believe_AIperception}	
\end{figure}

\begin{figure*}[ht]
	\centering
	\includegraphics[width=.99\textwidth]{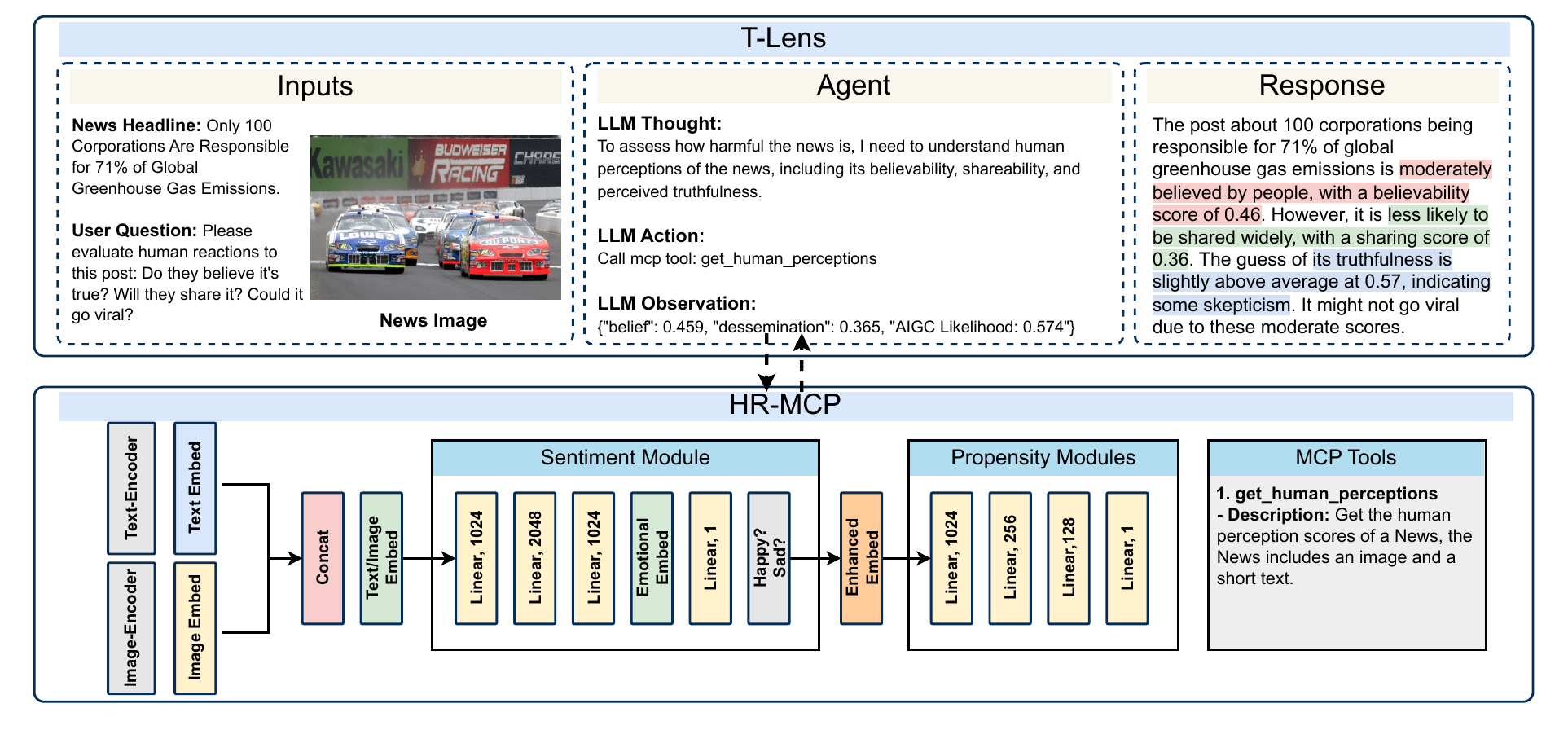}
	\caption[]{\textbf{T-Lens framework}: Predicting human responses through ReAct-style reasoning with integrated HR-MCP. Given multimodal inputs (text and image), the agent follows ReAct steps: \emph{Thought}, \emph{Action}, and \emph{Observation}. At each step, it determines whether human response signals (e.g., belief, dissemination tendency) are needed. If so, it queries \textbf{HR-MCP} (Human Response–Model Context Protocol), which returns predicted scores to guide the agent’s reasoning and generate explanations grounded in human reactions. \textbf{HR-MCP} is designed as a plug-and-play module, compatible with any LLM, making it easy to integrate into a wide range of agent systems.} \vspace{-0.15in}
	\label{fig:arch}
\end{figure*}

\noindent \emph{Observation 4:} As people believe information is more likely to be created by AI, their receptivity towards it decreases. However, well-crafted AIGC can still have a significant impact even when people perceive it to be AI-generated.

People were generally less likely to believe or share content perceived as AI-generated ($F$s $\geq 13.98$, $p$s $\leq .0001$; Fig. \ref{fig:sharing_believe_AIperception}). However, exceptions existed. We compared 50 highly likely AI-generated posts with high sharing intent (avg. share = 0.65) to 50 similarly perceived posts with low sharing (avg. = 0.05). Despite similar AI likelihood ($\sim 0.90$), the more shareable group had stronger text–visual consistency, emotional impact, and perceived trustworthiness (Fig. S14). This suggests that well-crafted AIGC can still gain traction, even when recognized as AI-generated.

\noindent \textbf{Discussion.} Our human study revealed relatively low human sensitivity to AIGC. This aligns with challenges identified in prior work on foundation models \cite{lu2023seeing,lin2024detecting} and highlights the importance of multi-modal consistency \cite{mink2022deepphish,kaate2023users}. Mismatched text and visuals often led to perceptions of AIGC, reducing belief and sharing intent \cite{bellaiche2023humans}. However, well-crafted AIGC showed high receptivity, even when users suspect its AI origin. These findings are based on current state-of-the-art models and may evolve with future AIGC quality. Our results are influenced by user backgrounds, with error bars representing human variance (see Suppl. Sec. 2 for analysis of individual differences). This human study contributes to discussions on AIGC's misinformation potential and inspired the development of T-Lens, our model that predicts human reactions to digital information.

\begin{table*}[t]
    \centering
    \small
    \resizebox{\textwidth}{!}{
    \begin{tabular}{l|cc|cc|cc|cc|cc|cc}
        \toprule
        \textbf{Model} & \multicolumn{2}{c|}{\textbf{AI likelihood}} & \multicolumn{2}{c|}{\textbf{Belief}} & \multicolumn{2}{c|}{\textbf{Dissemination}} & \multicolumn{2}{c|}{\textbf{Trustworthiness}} & \multicolumn{2}{c|}{\textbf{Impact}} & \multicolumn{2}{c}{\textbf{Openness}} \\
        & $\rho$ & AUC & $\rho$ & AUC & $\rho$ & AUC & $\rho$ & AUC & $\rho$ & AUC & $\rho$ & AUC \\
       \midrule
        
        \rowcolor{blue!20}
        \multicolumn{13}{c}{\textit{\textbf{Comparison with Conventional Models \& MLLMs}}} \\
        Eann \cite{wang2018eann} & 0.171 & 0.595 & 0.165 & 0.581 & 0.111 & 0.551 & 0.230 & 0.625 & 0.162 & 0.557 & 0.120 & 0.563 \\
        MCAN \cite{wu2021multimodal} & 0.219 & 0.600 & 0.260 & 0.635 & 0.198 & 0.604 & 0.226 & 0.620 & 0.248 & 0.604 & 0.277 & 0.609 \\
        CAFE \cite{chen2022cross} & 0.078 & 0.548 & 0.109 & 0.559 & 0.119 & 0.582 & 0.108 & 0.543 & 0.157 & 0.588 & 0.146 & 0.589 \\
        FNR \cite{ghorbanpour2023fnr} & 0.179 & 0.613 & 0.195 & 0.614 & 0.140 & 0.587 & 0.244 & 0.654 & 0.191 & 0.599 & 0.098 & 0.515 \\
 %       \midrule
        
        GPT-4o~\cite{openai2024gpt4o} & 0.314 & 0.679 & 0.333 & 0.667 & 0.132 & 0.622 & 0.353 & 0.685 & 0.251 & 0.595 & 0.183 & 0.600 \\
        Gemini-2.5-Flash~\cite{comanici2025gemini} & 0.337 & 0.689 & 0.340 & 0.667 & 0.098 & 0.562 & 0.371 & 0.749 & 0.252 & 0.636 & 0.108 & 0.564 \\
        Claude-3.7-Sonnet~\cite{claude3.7} & 0.245 & 0.621 & 0.276 & 0.645 & 0.164 & 0.658 & 0.310 & 0.600 & 0.235 & 0.581 & 0.182 & 0.590 \\
        \midrule
        
        \rowcolor{green!20}
        \multicolumn{13}{c}{\textit{\textbf{Ablation Study}}} \\
        Baseline: GPT-4o + HR-MCP (w/o visuals) & 0.145 & 0.575 & 0.186 & 0.603 & 0.179 & 0.601 & 0.195 & 0.724 & 0.230 & 0.621 & 0.181 & 0.590 \\
        Baseline: GPT-4o + HR-MCP (w/o text) & 0.278 & 0.634 & 0.233 & 0.620 & 0.185 & 0.609 & 0.317 & 0.682 & 0.205 & 0.594 & 0.164 & 0.577 \\
        Baseline: GPT-4o + HR-MCP (w/o sentiment) & 0.307 & 0.645 & 0.294 & 0.663 & 0.203 & 0.628 & 0.343 & 0.698 & 0.270 & 0.641 & 0.217 & 0.611 \\
        \textbf{T-Lens (Ours, GPT-4o + HR-MCP)} & \textbf{0.414} & \textbf{0.714} & \textbf{0.400} & \textbf{0.701} & \textbf{0.284} & \textbf{0.661} & \textbf{0.375} & \textbf{0.751} & \textbf{0.293} & \textbf{0.717} & \textbf{0.298} & \textbf{0.661} \\
        \bottomrule
    \end{tabular}
    }
    \caption{Merged results: Conventional models, LLMs, and ablation settings evaluated on six human response attributes with Spearman's $\rho$ and AUC.} %Missing values are indicated by "--".}
    \label{tab:merged_results}
\end{table*}

\section{4 \: Framework Design}

We introduce \textbf{T-Lens}, a ReAct-style agent system designed to reason not only over content but also over how humans are likely to perceive and respond to multimodal information (see Fig. \ref{fig:arch}). At the core of T-Lens is the \textbf{HR-MCP} (Human Response–Model Context Protocol), a dedicated MCP server trained to predict six key types of human responses to misinformation. Guided by insights from our human study, HR-MCP enables the agent to align its reasoning with how users interpret and emotionally react to multimodal content.

\subsection{Overview of T-Lens Framework}
T-Lens is built on the ReAct agent~\cite{yao2023react} and augmented with HR-MCP tools to support nuanced reasoning over user prompts and multimodal inputs (text and images). For each task, T-Lens performs up to 15 reasoning steps (configurable based on complexity), during which the LLM determines whether a human-like interpretation is needed. If so, it invokes one of the HR-MCP tools. Each step in T-Lens includes a \textit{thought} (rationale for calling a mcp tool), an \textit{action} (mcp tool name and parameters), and an \textit{observation} (mcp tool output). This design allows the agent to iteratively gather relevant semantic, emotional, and contextual cues before producing a final user-facing response. T-Lens supports interfaces with GPT-4o and Gemini, and is also compatible with open-source models like LLaMA~\cite{Touvron2023LLaMAOA}, Qwen~\cite{bai2023qwen}, and DeepSeek~\cite{liu2024deepseek}. In this paper, we use GPT-4o, one of the most proven and widely-used LLMs, as the agent's backbone. A detailed prompt example and a screenshot are shown in Figs.~S15 and S16 in the Supplementary.

\subsection{Design of HR-MCP}
HR-MCP is an MCP server integrated within T-Lens to model human responses to multimodal content. Since AIGC often combines text and images, we adopt a CLIP-style architecture with a ViT-based image encoder and a BERT-style text encoder. The model is trained on human-labeled data, making it well-aligned with identified human responses to multimodal information.

\noindent \textbf{Multimodal Semantics Consistency.} Through our human study, we found that discrepancies between text and visual significantly influenced the classification of content as AI-generated. Perceived content veracity emerged as the primary factor for AI news detection. Furthermore, the likelihood of content being AI-generated significantly impacted human receptivity. Drawing from these insights, the initial step of HR-MCP is to compute information semantics and model multi-modal semantic consistency. To achieve this, we use a Vision Transformer (ViT) \cite{dosovitskiy2020image} to obtain image embeddings and a BERT-like architecture with 12 layers of transformer blocks, each with a hidden size of 768 and 12 self-attention heads, to compute text embeddings. Specifically, we utilize the base version of ViT with a $32\times32$ patch size, taking a $224\times224\times3$ image as input and outputting a 512-dimensional image embedding. For text encoding, the encoder handles tokenized sentences, resulting in a 512-dimensional text embedding. Concatenating these produces a 1024-dimensional multi-modality embedding, encapsulating the semantics features from each modality as well as text-visual consistency. By such, HR-MCP learns key factors impacting AIGC perception and human receptivity, as identified in our human study.

\noindent \textbf{Multimodal Sentiment Consistency.} Our human study found that emotional mismatches between text and visuals indicate AIGC likelihood. Hence, after extracting multimodal embeddings, we add a sentiment module to compute emotion consistency between text and corresponding visuals. The module, an MLP-style subnetwork, takes a 1024-dimensional raw multi-modality embedding as input. The subnetwork consists of three linear layers (1024, 2048, and 1024 nodes), uses ReLU activation, and includes a dropout layer to encourage convergence and prevent overfitting. The sentiment module outputs a 1-dimensional feature representing sentiment consistency between text and visuals. 

\noindent \textbf{Embedding Fusion.} Obtaining two embeddings (semantics and sentiment) from previous steps, we sum them and apply Layer Norm for smoother gradients and convergence, yielding a 1024-dimensional enhanced embedding. This ultimate embedding encompasses image and text content, along with potential consistency between text and visuals in terms of both content and emotions.

The final submodule contains three identical propensity modules to predict three human responses (AIGC likelihood, belief, dissemination propensity) and compute three other metrics for human receptivity (trustworthiness, impact, and openness) based on the equations in Table \ref{tab:openness}. Each employs an MLP with four linear layers (1024, 256, 128, and 1 nodes), using ReLU activation and dropout after each layer.

\section{5 \: Experiments}

\subsection{5.1 \: Experimental Settings}

The HR-MCP model parameters for both the image encoder and text encoder are initialized using the pre-trained CLIP model \cite{radford2021learning}. Throughout the training process, we used a joint MSE loss of the predicted information attributes (AI likelihood, Belief, Dissemination). The parameters of the deep neural network model are learned in an end-to-end manner, leveraging the Adam optimizer with a learning rate of 0.0005. We trained the model for 50 epochs with a batch size of 64. The complete training procedure takes approximately 0.5 hour on a single NVIDIA 3090Ti GPU and the PyTorch framework. We randomly split the 3074 pieces of multimodal information into 80\% as a training set and 20\% as a test set. 

\noindent \textbf{Evaluation Metrics.} We used Spearman's correlation \((\rho)\), a standard metric for modeling human reactions~\cite{ICCV15_Khosla,singh2023long,kramer2023features}, and AUC to capture binary decision-making behavior (\emph{e.g.}, whether to share a post). Since decisions are often binary, we binarized annotations at a $0.5$ threshold; for example, we argue that a predicted $0.8$ better matches a ground truth of $0.6$ than $0.4$.

\vspace{0.015in}

\noindent \textbf{Evaluation Metrics.} As no existing models directly target human receptivity like T-Lens, we compare it with four multimodal misinformation detectors: FNR~\cite{ghorbanpour2023fnr}, CAFE~\cite{chen2022cross}, MCAN~\cite{wu2021multimodal}, and the widely used EANN~\cite{wang2018eann}. We exclude some recent models due to dataset or preprocessing limitations\footnote{E.g., SAFE~\cite{zhou2020similarity}, KDCN~\cite{sun2023inconsistent}, iMMoE~\cite{ying2023bootstrapping}, SNIFFER~\cite{qi2024sniffer}}. All models were trained with consistent five-fold cross-validation, and results are averaged over five runs. More importantly, we also compare T-Lens with recent powerful MLLMs, including GPT-4o~\cite{openai2024gpt4o}, Gemini 2.5 Flash~\cite{comanici2025gemini}, and Claude 3.7 Sonnet~\cite{claude3.7}, applied without additional tuning. This allows us to assess T-Lens against top-tier and widely-used MLLMs in modeling human receptivity.

\subsection{5.2 \: Experimental Results}

Table~\ref{tab:merged_results} showcases our method's robust performance on all metrics, surpassing comparison methods, MLLMs and all baselines. T-Lens excels in predicting participants' AIGC likelihood and belief ($\rho$s $\geq .400$, $AUC$s $\geq .701$) and effectively assesses information trustworthiness and impact ($\rho$s $\geq .375$, $AUC$s $\geq .717$), illustrating its potency in evaluating human receptivity. T-Lens is less effective in predicting dissemination (\(\rho = .284\), \(AUC = .661\)), likely due to the complex nature of sharing decisions, which are influenced by factors such as a user's background (\emph{e.g.}, political bias, cultural preferences, etc) and online behavior. Overall, our model demonstrates strong predictive power in both regression and binary classification tasks, showing promising ability in predicting human responses to digital information. The modest performance of the comparison methods underlines their design for disparate purposes, such as multimodal misinformation detection, rather than their comprehensive efficacy. The superior performance over MLLMs highlights the contribution of HR-MCP, which is specifically designed to model human responses to online information.

\vspace{0.01in}

\noindent \textbf{Ablation Study.} We compare our method against three baselines to evaluate the impact of our proposed modules on simulating human responses. The baselines include a model with only text, a model with only visuals, and a base T-Lens model without the sentiment module. As shown in Table \ref{tab:merged_results}, T-Lens outperforms all baselines in both regression and binary classification tasks, demonstrating the superiority of combining multi-modality. Notably, the sentiment module significantly bolsters performance, mirroring our human study's finding on the role of sentiment consistency. This finding highlights the importance of incorporating affective dimensions into computational models for a more comprehensive understanding of human-AI interactions. Our experimental results also align with previous research, which emphasizes that affective dimensions play a crucial role in misinformation \cite{ecker2022psychological}.

\vspace{0.01in}

In summary, our experiment suggests that, by incorporating both textual and visual, as well as the sentiment module, T-Lens effectively aligns with human cognitive processes, providing a more accurate simulation of how users interact with digital information. This comprehensive approach ensures that the model can better predict human responses, making it a valuable tool for combating misinformation in a multifaceted media landscape.

\noindent \textbf{Generalizability and Application.} To evaluate T-Lens' generalizability, we directly tested it without fine-tuning, on another dataset Mediaeval2015 \cite{boididou2018detection}, with 1022 multimodal tweets and their corresponding retweet counts---akin to dissemination propensity in our case. Our model achieved a \(\rho\) of $0.270$ in predicting retweet counts, close to the results on MhAIM, suggesting promising potential for T-Lens' applicability to real-world scenarios. 

\noindent \textbf{Discussion and Future Work.} Experiments show T-Lens's potential, particularly its ability to infer trustworthiness by prioritizing multimodal consistency over AIGC detection, aligning with human perception. Unlike models relying on external verification, which can be unreliable with the rise of AI-generated misinformation, T-Lens leverages insights from our human study, mirroring real-world behavior where people often don't cross-verify information. While T-Lens demonstrates promising results, future research could enhance its capabilities by incorporating individual behavioral traits that influence receptivity to misinformation, thus providing a more comprehensive understanding of the human-AI-information interaction landscape.

%\vspace{-.10in}
\section{6 \: Conclusion and Discussion}

In this work, we study and model human responses to AI-generated content to mitigate misinformation in its increasing prevalence. We introduce the MhAIM Dataset, comprising 111,153 pieces of AI-generated and 43,369 pieces of human-crafted multimodal information. We introduce T-Lens, an agent system which predicts how people will respond to multimodal information with reasoned explanations. The component of T-Lens is HR-MCP (Human Response Model Context Protocol), which uses the standardized Model Context Protocol, allowing it to seamlessly integrate into any LLM to enhance tasks that require modeling human responses. This research represents one of the first attempts to evaluate generative AI's misinformation potential from a human-centric, multimodal viewpoint.

\vspace{0.01in}

\noindent \textbf{Limitations.} Our study has several limitations, including the challenge of replicating real-world AIGC (mis)information creators and the fast-evolving nature of generative AI tools. Findings are based on a limited participant pool, with responses potentially influenced by individual backgrounds. While comprehensive, our dataset does not cover all content types or global perspectives, but provides a foundation for modeling human responses. Continued research is needed to keep pace with AI and behavioral shifts, including political bias and personal variance. We plan to update MhAIM with emerging AI trends, following practices in datasets like FVC \cite{papadopoulou2019corpus}.

\bibliography{arxiv}

\end{document}